\documentclass[runningheads]{llncs}
\usepackage{graphicx}
\usepackage[shortlabels]{enumitem}
\usepackage{url}
\usepackage{caption,subcaption}

\usepackage{adjustbox}
\usepackage{array}
\usepackage{multirow}
\usepackage{siunitx,booktabs}
\usepackage{algorithm}
\usepackage{algpseudocode}
\usepackage{makecell}
\usepackage{booktabs}
\usepackage{todonotes}
\usepackage{mathtools}
\usepackage{dsfont}
\newcolumntype{R}[2]{%
    >{\adjustbox{angle=#1,lap=\width-(#2)}\bgroup}%
    l%
    <{\egroup}%
}

\renewcommand{\epsilon}{\varepsilon}

\newcommand{\sobol}{Sobol\kern-0.15em\'{ } }

\usepackage{etoolbox}

\BeforeBeginEnvironment{figure*}{\vskip-2ex}
\AfterEndEnvironment{figure*}{\vskip-1ex}

\begin{document}
\title{Identifying Easy Instances to Improve Efficiency of ML Pipelines for Algorithm-Selection}

%
\titlerunning{Identifying Easy Instances to Improve Algorithm-Selection}
%
\author{Quentin Renau\orcidID{0000-0002-2487-981X} 
\and Emma Hart\orcidID{0000-0002-5405-4413}}
\authorrunning{Renau et. al.}
%
\institute{Edinburgh Napier University\\
\email{\{q.renau,  e.hart\}@napier.ac.uk}
}

\maketitle              
\begin{abstract}
Algorithm-selection (AS) methods are essential in order to obtain the best performance from a portfolio of solvers over large sets of instances. However, many AS methods rely on an analysis phase, e.g. where features are computed by sampling solutions and used as input in a machine-learning model. For AS to be efficient, it is therefore important that this analysis phase is not computationally expensive.  We propose a method for identifying \textit{easy} instances which can be solved quickly using a generalist solver without any need for algorithm-selection. This saves computational budget associated with feature-computation which can then be used elsewhere in an AS pipeline, e.g., enabling additional function evaluations on \textit{hard} problems.
Experiments on the BBOB dataset in two settings (batch and streaming) show that identifying easy instances  results in substantial savings in function evaluations. Re-allocating the saved budget to hard problems provides gains in performance compared to both the virtual best solver (VBS) computed with the original budget, the single best solver (SBS) and a trained algorithm-selector.

\end{abstract}

\section{Introduction}
\label{sec:intro}

For any large set of instances, it is well known that different algorithms will elicit different performances, resulting in the need to perform algorithm-selection in order to maximise performance. Typically this is achieved through the use of machine-learning (ML)  methods which either predict the label of the best solver or predict the performance of  an algorithm ~\cite{kerschke_automated_2019}. In order to apply an algorithm-selector, some computation is usually required to obtain a feature-vector used by the ML model.
In continuous optimisation, the most common approach is to compute Exploratory Landscape Analysis (ELA) features~\cite{mersmann_exploratory_2011}, while recent alternatives use short \textit{probing trajectories} as the model input, obtained by running one or more solvers on an instance~\cite{kostovska2022per,RenauH24}. 
However, regardless of the type of input chosen, it is crucial to ensure that the analysis required to compute the input data required by a selector does not itself become too expensive~\cite{KotthoffKHT15}.

One obvious way to reduce the effort spent on computing the information required to train an algorithm-selector would be to identify \textit{easy} problems that can be solved to a desired level of precision by a good general solver without any need for a selection process. This is particularly relevant given that Kerschke {\em et. al.}~\cite{kerschke2019automated} note that algorithm selection systems generally perform poorly on easy function instances. Furthermore, an open-issue in continuous optimisation is that ``some problems are embarrassingly easy (e.g., sphere, linear slope)"~\cite{ELA23}, leading to the authors asking ``is there a (cheap) way to distinguish easy from complex problems?".
The same sentiment is echoed in~\cite{figueira2017easy} in the context of multi-objective optimisation where the authors note that it is ``easy to say they [instances] are hard, but hard to say they are easy".  While there is a wealth of literature directed at identifying hard problems, particularly in combinatorial optimisation~\cite{smith2010understanding,smith2021revisiting}, there is very little literature directed towards identifying easy ones.

In response to the issues just outlined, we propose a revision to the algorithm-selection pipeline which includes an additional step to predict whether an instance is easy or hard prior to potentially applying a selector. This additional model acts as a filter for recognising  `easy' instances which can be directly solved by a general solver with no need to apply algorithm-selection.   The information needed to  detect whether an instance is easy is a subset of the information needed by the algorithm-selector: it can be cheaply computed and reused in the selection step if needed.
In addition, running the general solver on an easy instance can often produce a solution of desired quality with less budget than the maximum allocated. The unused budget can then  be \textit{re-allocated} to enable additional function evaluations on `hard' instances. 

We use the BBOB benchmark test-suite as a testbed~\cite{bbob-functions} to test the new pipeline. Its performance is evaluated in two settings. In the typical `batch' setting, a set of instances need to be solved and are all available when the solving process starts. 
We also consider a `streaming' setting which is typical in many real-word applications: here an infinite stream of instances arrive either periodically or sporadically~\cite{degorre2008scheduling}, and each must be solved on arrival, with no knowledge of what might arrive downstream in the future. Specific examples appear in many domains, ranging from
the allocation of machines in a factory~\cite{pinedo2005planning}, allocation of processor time slots in a real-time system~\cite{buttazzo2011hard}, allocating communication channels in a network~\cite{gan2007scheduling} or allocation of vehicles for transportation tasks~\cite{raff1983routing}. 

We demonstrate that:
\begin{itemize}
    \item Without budget re-allocation, including a classifier to identify easy instances in a pipeline results in a gain in performance compared to using a single trained algorithm-selector. Furthermore, there is only a small loss in performance when compared to the virtual best solver (VBS).
    \item When re-allocating budget saved by identifying easy instances, we enable extra function evaluations to be performed on hard instances. This results in a considerable performance gain for both the batch and streaming setting, even in comparison to the VBS (calculated using a fixed evaluation budget per instance). 
\end{itemize}

\paragraph{Reproducibility:} Code and data related to this study can be found at~\cite{dataHardness}.

\section{Related Work}
\label{sec:related}

Kothoff {\em et. al.}~\cite{KotthoffKHT15} note that although algorithm-selectors can be beneficial, it is important to ensure that the computational cost of any analysis needed to inform the selector is minimised, stating `if selecting an algorithm for solving a problem is more expensive than solving the problem, there is no point in doing so'. 
Note that the cost of using a selector refers to both the cost of computing the input and of executing the selector. In the domain of Boolean satisfiability (SAT) solving, \cite{xu2008satzilla} propose the use of a \textit{pre-solver} --- an algorithm with good general performance --- and start solving an instance with this while simultaneously analysing the instance.
The idea is that easy instances will be solved before the analysis phase finishes therefore  mitigating the need for an additional algorithm-selection step.
However, Tanabe~\cite{Tanabe22} notes that no previous study has used a pre-solver for black-box optimisation so it is unclear to what extent the same process could be used.
However, this literature underpins the argument that recognising easy instances in order to save computational effort is an important line of research for the field.  
Unfortunately, it is equally clear that recognising easy problems is not easy:  \cite{figueira2017easy} discuss this in the  context of multi-objective combinatorial optimisation problems while~\cite{ehrgott2000survey} conclude that it is ‘hard to say it's easy' in general (in relation to combinatorial optimisation).

In the continuous optimisation domain, the vast majority of previous work in AS relies on ELA features which are not cheap to compute: the recommended budget is usually $50d$ samples, with $d$ the dimension of the problem. Some recent work tries to reduce the cost of feature computation by extracting ELA features from a trajectory obtained by running a solver of interest, rather than computing ELA features on a set of  separate samples that are then discarded~\cite{JankovicVKNED22,JankovicED21}. 
Similarly, in~\cite{CenikjPDKE23}, statistics derived from a fitness trajectory are used as features for a selector, however, the budget used to calculate the trajectory is $6$ times more than the recommended ELA features budget. In~\cite{RenauH24}, each solver in a portfolio is run for a short time to create a `probing-trajectory', i.e., short fitness trajectory and this data is used to directly obtain a prediction from a selector: as the selected solver can then be run starting from the point the probing-trajectory terminated, thus saving some budget although this is never re-used. \cite{renau2024improving} pushes further in this direction in using a selector that uses very short trajectories derived from running a simulated annealing algorithm that has been tuned to generate discriminating trajectories. Despite these recent efforts to reduce the cost of feature-computation, to the best of our knowledge, we are unaware of any cheap method to identify easy instances which can be solved by a generalist solver instead of passing them through an algorithm-selector and can reuse any saved budget to better solve future instances.

\section{Motivation}
\label{sec:pipeline}

\begin{figure*}
    \centering
    \includegraphics[width=\textwidth]{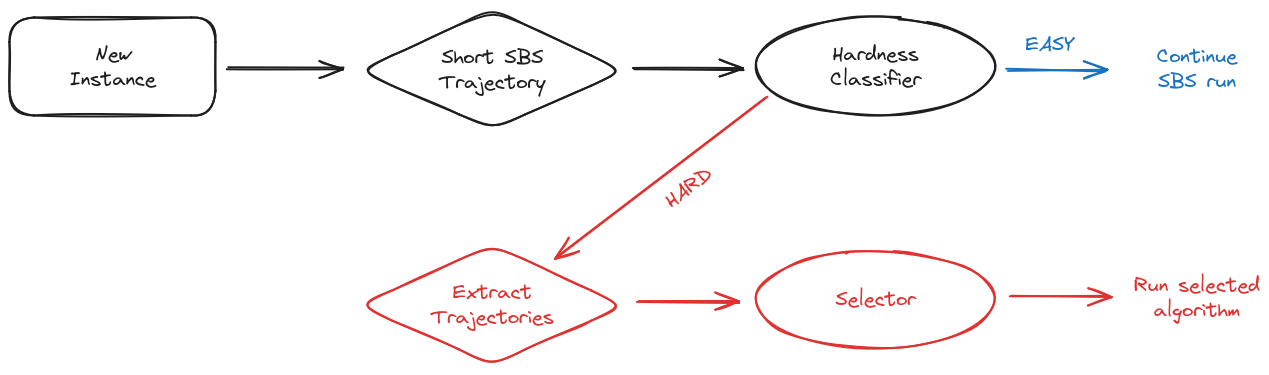}
    \caption{Algorithm-selection pipeline including the `easy' instance filter using short trajectories.}
    \label{fig:pipeline}
\end{figure*}

Fig.~\ref{fig:pipeline} provides an overview of the proposed pipeline in which a new step is introduced to detect whether an instance is easy to solve.  First, a small budget  $\varphi_h$ is used to obtain a feature vector which is passed to the `hardness classifier': if the instance is classified as easy then the single best solver (SBS) from the portfolio is used to solve the instance with budget $b_h$.  If an instance is labelled as hard, then algorithm-selection proceeds as normal: further input-data for the selector is calculated with budget $\varphi_{AS} > \varphi_h$, and the selector outputs the label of the best solver from the portfolio which is solved using budget
$b_{AS} \geq b_h$.

One of the proposed advantages of the pipeline is that it offers an opportunity to dynamically re-allocate budget saved from recognising easy instances to extend the solving budget for hard instances. 
As additional budget is required to generate features for the algorithm-selector, then this extra budget is saved every time an instance is classified as easy.  In addition, a lower budget $b_h$ can be allocated for \textit{solving} instances identified as easy, freeing up further budget. Alternatively, the SBS used to solve easy instances can be terminated once a desired level of precision is reached, i.e. at budget $b_t$, hence again freeing budget --- the amount of budget saved each time will vary depending on how quickly the desired level of precision is reached. Therefore, the total budget saved per easy instance is either $\varphi_{AS} + (b_{AS} - b_h)$ or $\varphi_{AS} + (b_{AS} - b_t)$, depending on whether the run is terminated before $b_h$ evaluations. 

The saved budget is re-allocated among hard instances in a manner that depends on the setting:

\begin{itemize}
    \item In the batch setting in which all instances are known at the start, the \textit{total} budget  saved  $B$ from solving easy instances is divided equally amongst the remaining $n_h$ hard instances, thus extending the solving budget for each instance by $B/n_h$
    \item In the streaming setting, the budget $b_i$ saved on an easy instance is immediately allocated in its entirety to the \textit{next} hard instance that arrives in the stream to extend the run of the predicted solver. 
\end{itemize}
The details of how each step is instantiated in the pipeline are provided in the next section.

\section{Methods}
\label{sec:methods}
We use the Black-Box Optimisation Benchmark (BBOB) functions from the COCO platform~\cite{cocoJournal} as a test-bed. In particular, we consider the first $5$ instances of the $24$ noiseless functions in dimension $d=10$.
Instances are variations of the same base function such as rotations, translations or scaling.
We consider a portfolio of three algorithms --- CMA-ES~\cite{HansenO01}, Particle Swarm Optimisation (PSO)~\cite{KennedyPSO95}, and Differential Evolution (DE)~\cite{StornP97}. In order to label each instance with the best solver, each algorithm is run $5$ times per instance for $4{,}000$ evaluations, and the median fitness is recorded. CMA-ES is best on $11$ functions, PSO on $7$, and DE on $6$. CMA-ES is therefore designated as the single best solver (SBS), i.e.,  the algorithm having the best median performance.  This data is obtained directly from~\cite{dataDiederick} which records search trajectories per run and is described in detail in~\cite{Vermetten2022}. 
Note that as these authors performed some tuning, each solver has a different population size: these are ($10$,$30$,$40$) respectively for CMA-ES, DE and PSO.

\subsection{Hardness Classifier}
\label{sec:hardness}
In previous work~\cite{RenauH24,renau2024improving}, we demonstrated that training an algorithm-selector using \textit{probing-trajectories} --- a short time-series of fitness information obtained from running one or more solvers --- outperformed classifiers trained using ELA features on the BBOB suite. Therefore we also utilise this approach here to  provide input to a `hardness' classifier. This has the advantage that if an instance is detected as easy, then an instance is simply continued from where the probing trajectory terminated. We label a function as \textit{easy} if \textit{all} $100$ runs of the SBS have a performance below $10^{-7}$, and \textit{hard} otherwise. 
This threshold has been chosen arbitrarily and represents a solution that may not be optimal but good enough in practice.
For most problems, practitioners are usually able to set a threshold representing a good enough solution.
Note that the data is imbalanced: only $3$ of the $24$ functions are labelled as easy\footnote{F1, F5, and F6}. 

The hardness classifier is trained using a single short trajectory obtained by running the single best solver (CMA-ES) for a short time ($7$ generations with a population size $10$)\footnote{$7$ is chosen as this results in a total number of function evaluations that is close to the recommended ELA budget of $500$}. We use an LSTM network~\cite{lstm} as classifier. An LSTM is a type of recurrent network that is able to keep track of arbitrary long-term dependencies in input sequences. Given that our input is a time-series, this is a natural choice. LSTMs have previously been demonstrated to be useful in classifying time-varying data in the online bin-packing domain~\cite{AlissaSH23}. The LSTM is trained for $200$ epochs, with a learning rate of $0.0001$ and optimised with Adam~\cite{kingma2014adam}.  These parameters were found through empirical tuning. We provide an implementation of the network that can be found in~\cite{dataHardness}. The model outputs a binary classification that labels an instance as `easy' or `hard'. 

To train and evaluate the hardness classifier,  we use a $5$-fold validation procedure in which a randomly selected 80\% of the instance-data is used for training and the remainder for testing, repeated $5$ times. 
The total number of samples  is $24\times 5\times 100 = 12{,}000$, i.e $24$ functions, $5$ instances and with $100$ trajectories per instance. Each test-set contains $20\%$ of this data, i.e. $2{,}400$ instances which are presented one by one to the hardness classifier. The model used in the final pipeline is the model with median accuracy from each of the $5$ splits.

\subsection{Algorithm-Selection}
\label{sec:algsel}

When the output is `hard', the hardness classifier redirects the pipeline to a classical algorithm-selection model. The input to this classifier consists of concatenated probing-trajectories from each solver in the portfolio. This concatenated vector was demonstrated in~\cite{RenauH24} to provide improved performance compared to training a classifier with a single probing-trajectory from one solver. Note that the probing-trajectory for the SBS has already been calculated in order to run the first part of the pipeline, so two additional trajectories (from PSO and DE) need to be computed to obtain the input required for this classifier.

To train the model we use trajectories obtained over $7$ generations of each of the three algorithms (which corresponds to $7\times (10+30+40) = 560$ function evaluations in total for the portfolio)\footnote{Recall from Section~\ref{sec:methods} that each of three solvers have different population sizes in the data provided by~\cite{dataDiederick}}.
The algorithm-selection procedure is a classification task, i.e., given a concatenated trajectory representing the instance at-hand, the selector outputs the label of the algorithm that should be run to solve the instance.
We compile a dataset using data directly obtained from~\cite{dataDiederick}. This contains $5$ runs of each solver per instance. We create a concatenated trajectory per instance per run, resulting in $(24\times 5 \times 5) = 600$ trajectories.
Each trajectory is labelled with the solver that produced the best median performance over the $5$ runs of the $5$ instances.
The algorithm-selector is also an LSTM as described in Section~\ref{sec:hardness} and is trained in exactly the same manner using a $5$-fold validation procedure. We test each of the $5$ trained models in turn in the pipeline. 
For the batch scenario, the instances in each test set can be presented in any order; for the streaming scenario, each instance is presented to the pipeline in exactly the order it occurs in the test set.

\subsection{Saving and Re-allocating budget}
We investigate different methods of re-allocating saved budget, depending on the scenario setting (batch or streaming) as described in Section~\ref{sec:pipeline}. We propose two ways to save budget, which can be used in both scenarios:

\begin{itemize}
    \item \textit{Save Selector Budget (SSB)}: if an instance is classified as easy using the SBS trajectory, then there is no need to obtain trajectories for the other solvers in the portfolio that are needed to use the algorithm-selector. In this case, the saved budget $b_{SSB}$ is equal to $7 \times (30+40) = 490$ ($7$ generation trajectories not extracted from DE and PSO).
    
    \item \textit{Save Selector Budget and Curtail Easy Runs (SSB-CE)}: if an instance is classified as easy, then the budget for running the SBS is cut by a fixed amount, from $4{,}000$ function evaluations to $2{,}700$ (minimum budget needed for the SBS to reach the $10^{-7}$ performance for all runs on the three `easy' functions), therefore saving $1{,}300$ function evaluations on top of the $490$ from not evaluating the selector.
\end{itemize}

In the \textit{batch} setting, the budget saved \textit{per instance} is accumulated over the entire batch of instances. If there are $n_h$ hard instances, this budget is then divided equally between each instance to extend the run of each solver predicted by the algorithm-selector. In the \textit{streaming} setting, budget saved on an easy instance is re-used according to a na\"{i}ve strategy that extends the length of the run of the \textit{next} instance to be identified as hard with the full saved budget from the previous step. This strategy is simple but provides a reasonable baseline. We discuss potential alternatives in Section~\ref{sec:conclusion}.

\section{Results}
We first establish the accuracy of the hardness classifier independently of the pipeline and then propose some baselines in which there is no re-allocation of budget. Under this setting (i.e., no re-allocation), both the batch and streaming scenarios are equivalent. Following this, we evaluate the full pipeline on both batch and streaming scenarios using the re-allocation strategies outlined above.

\subsection{Baselines}

\begin{figure}
    \centering
    \includegraphics[width=0.4\textwidth]{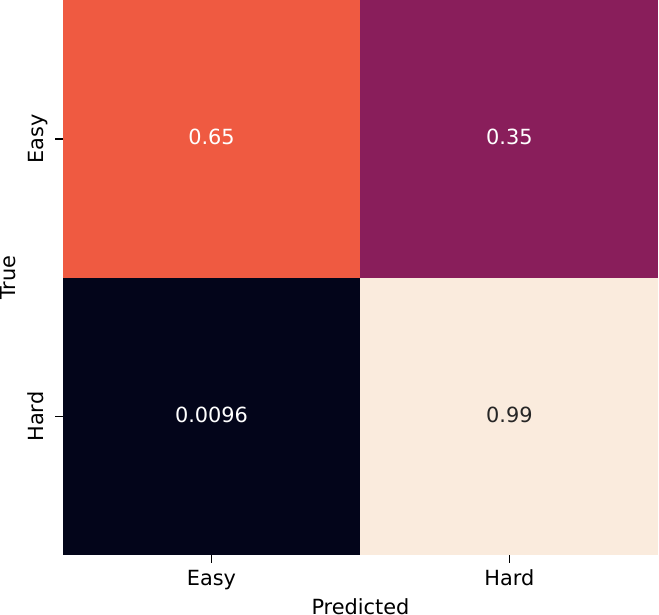}
    \caption{Median confusion matrix of the hardness prediction.}
    \label{fig:cm}
\end{figure}

Fig.~\ref{fig:cm} shows the median confusion matrix of the $5$ random sub-sampling splits for the hardness classifier. The median accuracy score is $94.5\%$ and the median balanced accuracy score (accounting for imbalanced data) is $82.1\%$.
The imbalance in the data mentioned previously is reflected in the confusion matrix. The `hard' class is over-represented in the data and is easily classified with $99\%$ accuracy. On the other hand, the under-represented `easy' class has an accuracy of $65\%$. This is not unexpected as we did not implement any mechanisms to rectify the imbalance such as weighting classes or oversampling. Nevertheless, the hardness predictor is still useful as discussed below. 

Firstly, even if `easy' instances are only recognised $2/3$ of the time, budget is saved on each of these instances.
This could amount to a significant saving: for \textit{each} instance identified as easy we save  $((30 \times 7) + (40 \times 7))+1{,}300 = 1{,}790$ evaluations in the SSB-CE setting: i.e., the number of evaluation needed to compute the PSO and DE trajectories for input to the AS which are run for population sizes ($30$,$40$) respectively for 7 generations, and the $1{,}300$ evaluations saved by curtailing the budget.
For easy instances that are incorrectly classified as `hard', no additional cost is incurred --- without the additional check in the pipeline, these instances would have been passed to the algorithm-selector regardless.
Secondly, note from Fig.~\ref{fig:cm} that almost no `hard' instance is ever mistaken for an `easy' instance --- the accuracy on these instances is $99\%$.  Mistaking a `hard' instance for an `easy' one could lead to a significant loss in solution quality as the algorithm-selection step would be bypassed. On the contrary, mistaking easy instances as `hard' ones does not have any impact on the performance of the pipeline, although they represent a missed opportunity to save budget.

\begin{table}[h]
\caption{Sum of the loss to the VBS for each split of the batch setting. Results are shown for the SBS and the pipeline with and without re-allocation strategies. Positive values indicate a gain over the VBS. Bold indicates the best performance for each column.  The selector in the pipeline is the VBS selector.}
\label{tab:batch_vbs}
\centering

\begin{tabular}{c|c|cccccc}
\hline
\multicolumn{1}{l|}{}          & Method              & \begin{tabular}[c]{@{}c@{}}Overall\\ loss to\\  VBS\end{tabular} & \begin{tabular}[c]{@{}c@{}}Loss to\\ VBS\\  split1\end{tabular} & \begin{tabular}[c]{@{}c@{}}Loss to\\ VBS\\  split2\end{tabular} & \begin{tabular}[c]{@{}c@{}}Loss to\\ VBS\\  split3\end{tabular} & \begin{tabular}[c]{@{}c@{}}Loss to\\ VBS\\  split4\end{tabular} & \begin{tabular}[c]{@{}c@{}}Loss to\\ VBS\\  split5\end{tabular} \\ \hline
\multirow{2}{*}{Baselines}      & SBS                 & -15623.25                                                        & -3122.95                                                        & -2992.35                                                        & -2698.8                                                         & -3165.53                                                        & -3283.62                                                        \\
                               & Pipeline No Savings & -70.27                                                           & -18.05                                                          & -6.27                                                           & -24.32                                                          & -16.26                                                          & -5.38                                                           \\ \hline
\multirow{2}{*}{Re-allocation} & Pipeline SSB        & 129.68                                                           & \textbf{19.61}                                                  & 43.56                                                           & 16.53                                                           & 16.62                                                           & 33.36                                                           \\
                               & Pipeline SSB-CE     & \textbf{315.37}                                                  & -62.83                                                          & \textbf{67.07}                                                  & \textbf{122.89}                                                 & \textbf{92.17}                                                  & \textbf{96.07}                                                  \\ \hline
\end{tabular}
\end{table}

\begin{table}[h]
\centering
\caption{Sum of the loss to the VBS for each split of the batch setting. Results are shown for the SBS, the trained selector and the pipeline with and without re-allocation strategies. Bold indicates the lowest loss to the VBS for each column. The selector in the pipeline is a trained selector.}
\label{tab:batch_selector}

\begin{tabular}{c|c|cccccc}
\hline
\multicolumn{1}{l|}{}          & Methods             & \begin{tabular}[c]{@{}c@{}}Overall\\ loss to\\ VBS\end{tabular} & \begin{tabular}[c]{@{}c@{}}Loss to\\ VBS\\ split1\end{tabular} & \begin{tabular}[c]{@{}c@{}}Loss to\\ VBS\\ split2\end{tabular} & \begin{tabular}[c]{@{}c@{}}Loss to\\ VBS\\ split3\end{tabular} & \begin{tabular}[c]{@{}c@{}}Loss to\\ VBS\\ split4\end{tabular} & \begin{tabular}[c]{@{}c@{}}Loss to\\ VBS\\ split5\end{tabular} \\ \hline
\multirow{3}{*}{Baselines}     & SBS                 & -1373.93                                                        & -335.62                                                        & -251.15                                                        & -328.59                                                        & -276.23                                                        & -182.34                                                        \\
                               & Trained Selector    & -471.44                                                         & -53.07                                                         & -117.81                                                        & -116.11                                                        & -125.42                                                        & -59.04                                                         \\
                               & Pipeline No Savings & -466.14                                                         & -53.96                                                         & -117.81                                                        & -116.11                                                        & -123.34                                                        & -54.91                                                         \\ \hline
\multirow{2}{*}{Re-allocation} & Pipeline SSB        & \multicolumn{1}{l}{-450.78}                                     & \multicolumn{1}{l}{-51.15}                                     & \multicolumn{1}{l}{-115.27}                                    & \multicolumn{1}{l}{-114.35}                                    & \multicolumn{1}{l}{\textbf{-119.4}}                            & \multicolumn{1}{l}{-50.62}                                     \\
                               & Pipeline SSB-CE     & \multicolumn{1}{l}{\textbf{-438.67}}                            & \multicolumn{1}{l}{\textbf{-46.14}}                            & \multicolumn{1}{l}{\textbf{-110.09}}                           & \multicolumn{1}{l}{\textbf{-110.43}}                           & \multicolumn{1}{l}{-128.5}                                     & \multicolumn{1}{l}{\textbf{-43.51}}                            \\ \hline
\end{tabular}
\end{table}

To establish a baseline, we measure the accumulated loss with respect to the VBS (calculated using the fixed evaluation budget per instance of $4{,}000$) of (1) SBS; (2) the trained algorithm-selector only; (3) using the full pipeline that includes the hardness classifier but uses a `perfect' algorithm-selector, i.e. the VBS; (4) using the full pipeline that includes the hardness classifier and a trained model as the selector.
The loss is calculated as the sum of the performance $VBS_i - t_i$, where $i$ is an instance and $t$ is one of the four methods just described. The sum is calculated over each instance in each of the $5$ test sets for each of the four methods $t$.

The results are shown in Table~\ref{tab:batch_vbs} when the VBS is used as selector in the pipeline and  in Table~\ref{tab:batch_selector} when the trained selector is used in the pipeline ('Baselines' rows). Table~\ref{tab:batch_vbs} shows that the pipeline with no re-allocation outperforms the SBS but has lower performance than the VBS.
Both results are expected: since the selector in the pipeline is the VBS selector, it is expected to outperform the SBS.
Moreover, the pipeline adds an imperfect hardness classifier to the VBS selector, we expect to see a degradation compared to purely the VBS.

We observe in Table~\ref{tab:batch_selector} (Baselines rows) that both the trained selector and the pipeline without re-allocation outperform the SBS.
This result was expected given that no one solver outperforms the others on all functions and thus there is a benefit to performing algorithm-selection.
We also observe that both the trained selector and the pipeline with no re-allocation obtain similar losses to the VBS.
These results could also be anticipated since no budget is saved when the hardness classifier is used.
Interestingly, the performance of the pipeline is slightly better than the performance obtained by using the trained selector only, even though it only adds a hardness classifier.
The use of the hardness classifier seems to mitigate the selector mistakes, i.e., when the selector gets the algorithm-selection wrong on `easy' instances, then the use of the SBS as a result of classifying the instance as easy almost always leads to better results.

Overall, we observe that using the pipeline presented in Fig.~\ref{fig:pipeline} does not degrade the performances compared to a classical algorithm-selection approach.
The next sections present results where the pipeline is fully leveraged, i.e., where unused budget is saved and re-allocated.

\subsection{Re-allocating budget: Batch setting}
\label{sec:batch}
\paragraph{Algorithm Selection with VBS:}
We use a test set composed of instances that were not used to train the algorithm selector model, i.e., a batch is composed of $20\% \times 12{,}000 = 2{,}400$ instances.
The process is repeated five times, i.e. for each split.
These instances are first fed to the hardness classifier and the number of instances classified as `easy' counted. This determines how many extra function evaluations can be allocated to extending the runs for instances classified as hard. In the five repetitions of the batch settings, there are between $6.5\%$ and $10\%$ of instances are identified as  `easy'. This results in a median $49$ function evaluations added to each of the remaining instances classified as hard when using the SSB strategy and $179$ function evaluations when using the SSB-CE strategy.

Table~\ref{tab:batch_vbs} shows the overall results and results for each split of the data.
Interestingly, both re-allocation budget approaches outperform the VBS, i.e., re-allocating budget to harder instances improves the overall performance. The
SSB-CE  strategy obtains the best performance: its overall gain is more than twice the gain of SSB.
Moreover, on split $3$, SSB-CE obtains almost the same gain as SSB on the five splits combined.
However, SSB-CE can also be the worst-performing approach: on split $1$, the loss to the VBS is substantial and nearly matches the overall loss of the pipeline without savings.

\paragraph{Algorithm Selection with trained selector:}

In these experiments, we use exactly the same procedure outlined in the previous paragraph but replace the VBS selector with a trained selector. Over the five repetitions, between $6.7\%$ and $10\%$ of instances are identified as `easy'. This results in a median $39$ function evaluations added to each of the remaining instances using the SSB strategy and $145$ function evaluations using the SSB-CE strategy.

Table~\ref{tab:batch_selector} shows the loss to the VBS for the three pipelines: without savings, SSB, and SSB-CE.
Even though the number of identified `easy' instances represents a small part of the whole batch and relatively few evaluations are added to the remaining functions, we observe a gain compared to the `no savings' result.
The best gain is obtained using the SSB-CE strategy with $5.9\%$ improvement compared to no savings.
Although this is the best-performing strategy overall, we should note that SSB-CE is outperformed by the trained selector in the split $4$. This is due to the curtailing of easy runs, i.e., running the SBS with a smaller number of function evaluations can lead to decreasing performances for instances  mistakenly labelled as `easy'.
The SSB strategy offer a good compromise, i.e., it performs $3.3\%$ better than the pipeline with no savings but consistently performs better when considered over all splits.

Overall, SSB-CE obtains better performances with both the VBS and trained selector but is also the most volatile strategy, i.e., although big gains are achieved overall, losses to classical algorithm-selection can arise in some settings.

\subsection{Re-allocating budget: Streaming setting}
Contrary to the batch setting, in this section we consider a stream where instances to solve arrive one by one and must be solved in the order they arrive.
We consider two pipelines: one where the algorithm selector is the VBS selector and one where the selector is a trained classifier.

\begin{figure*}
    \centering
    \includegraphics[width=\textwidth]{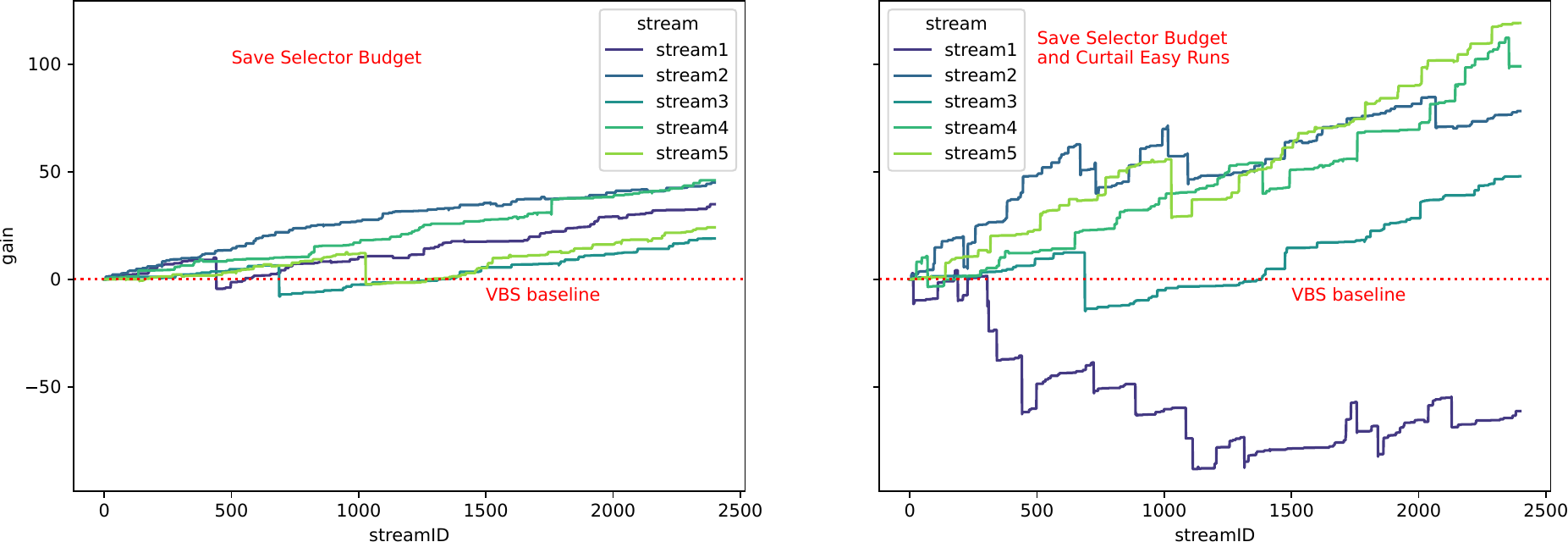}
    \caption{Cumulative difference between the full pipeline using the hardness classification and VBS selector when budget is saved and re-allocated.}
    \label{fig:budget}
\end{figure*}

\paragraph{Algorithm Selection with VBS:}
Fig.~\ref{fig:budget} displays the cumulative fitness gain obtained for each of the $5$ streams of unseen instances when budget is saved on `easy' instances and then reused on the next predicted instance that is predicted to be`hard'. The left figure shows the gain when only the budget from the selector is saved (SSB) while the right figure shows the results when easy runs are also curtailed,  i.e., fewer evaluations are performed on `easy' instances (SSB-CE).

Overall, for every model split, we observe a positive gain by using the extra budget on `hard' instances. The median gain in fitness value on SSB is $34.94$ while saving additional evaluations on the `easy' instances (SSB-CE) leads to a $78.22$ median gain, more than twice the gain obtained with the selector budget alone.
Despite these gains, we also observe losses for some instances. These occur due to a misclassification by the hardness classifier.
This has a bigger impact when using the SSB-CE strategy: if an instance is not easy and in particular if CMA-ES (the SBS) performs poorly on that function, then this results in a larger loss due to the curtailed run. This appears to have a particular impact in one of the five models (stream $5$).

\begin{figure}
    \centering
    \includegraphics[width=0.6\textwidth]{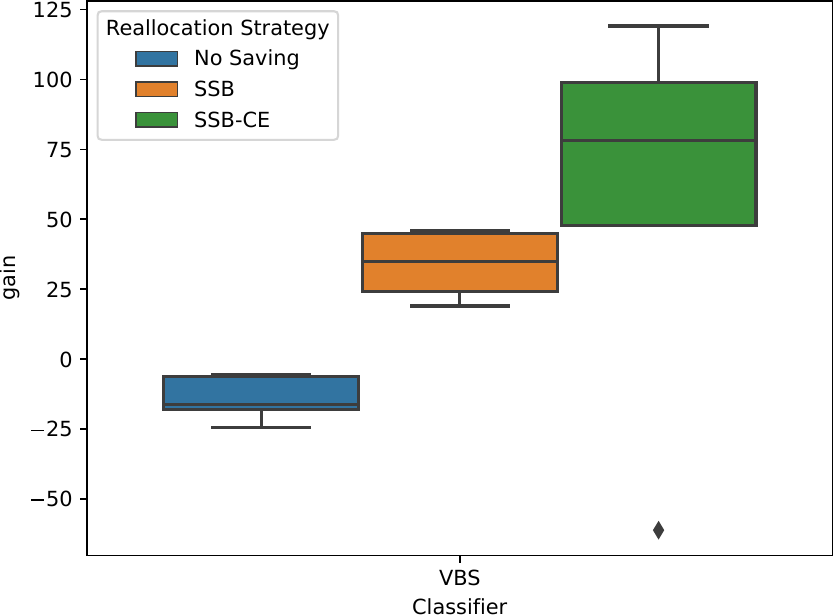}
    \caption{Boxplots of gains at the end of the stream of instances for each saving budget strategy with VBS selector.}
    \label{fig:boxplot_vbs}
\end{figure}

The results are summarised in Fig.~\ref{fig:boxplot_vbs}. This shows a small loss in comparison to the VBS when not re-allocating saved budget, contrasting to postive gains when using the two forms of budget saving and re-allocation. 
Here it is clear that by re-allocating unused budget, we realise a gain in performance compared to VBS performance (calculated as previously explained in Section~\ref{sec:algsel}). This is possible because re-allocating the budget enables longer runs on some instances, which then reach better fitness. As expected, this is maximised using the curtailing of easy runs which re-allocates more budget.

\paragraph{Algorithm Selection with trained selector:}

Recall from Section~\ref{sec:algsel} that the test datasets used to evaluate the $5$ trained algorithm-selectors models each only contain $120$ instances. In addition to this only being a short stream, it is important to recognise that any gain due to re-allocating saved budget will be influenced by the order in which instances are presented: this results from the simple strategy used which re-allocates saved budget to the next hard instance. Therefore, in order to quantify the effect of this, we create a longer stream in which we randomly select $120$ samples (with replacement) from the original $120$ test instances, repeating this $20$ times, resulting in a stream of length $2{,}400$. Each group of $120$ samples therefore may have a different set of instances and a different ordering.

\begin{figure}
    \centering
    \includegraphics[width=\textwidth]{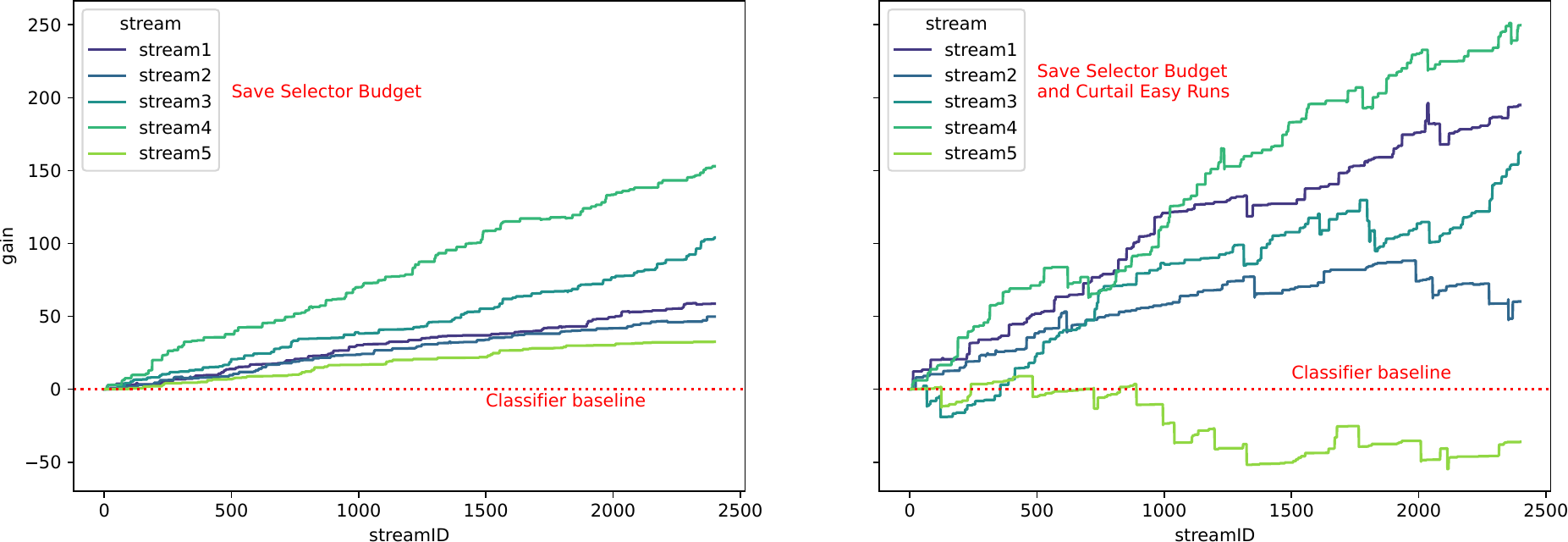}
    \caption{Cumulative difference between the full pipeline using the hardness classification and trained selector when budget is saved and re-allocated.}
    \label{fig:budget_NN}
\end{figure}

Fig.~\ref{fig:budget_NN} displays the cumulative fitness gains obtained on the $5$ streams of unseen instances when budget is saved on `easy' instances and then reused on the next predicted `hard' instance with a trained selector.
Overall, the same behaviour can be seen as observed with a VBS selector, i.e., saving budget improves the performance of the pipeline while saving budget by curtailing runs can lead to bigger gains but also to some losses in performance.
For every model but one, the gains obtained with SSB-CE are greater than using SSB. The latter provides a gain which is greater than only using the pipeline without budget re-allocation.
For example, for stream $4$, the gain of $75.5$ obtained by using the pipeline reaches $153$ using SSB and $249.7$ when SSB-CE is used.

\begin{figure}
    \centering
    \includegraphics[width=0.6\textwidth]{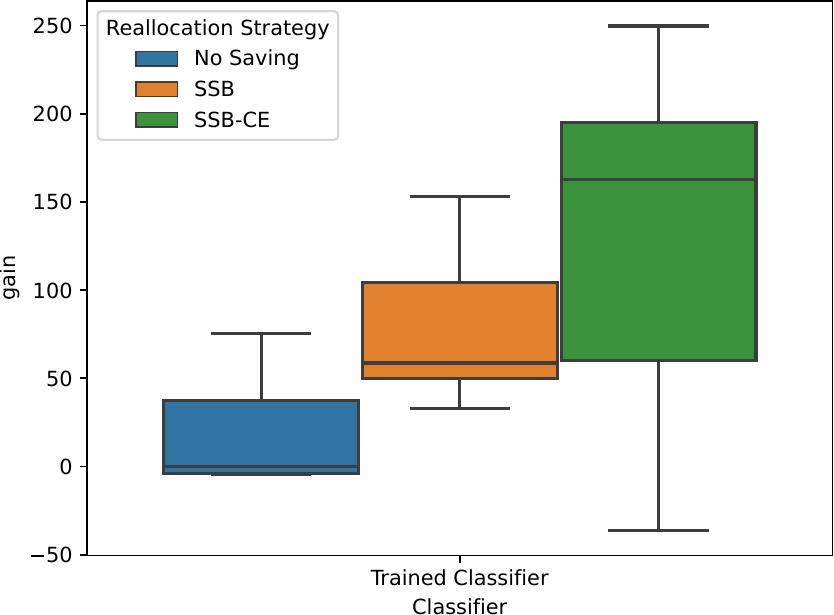}
    \caption{Boxplots of gains at the end of the stream of instances for each saving budget strategy with the trained selector.}
    \label{fig:boxplot_NN}
\end{figure}

Fig.~\ref{fig:boxplot_NN} summarises the performance gains using the pipeline that includes the hardness classifier compared to an algorithm-selector only. 
As described in Section~\ref{sec:batch}, we see a small gain, even when no budget re-allocation mechanism is used. Much larger gains are seen by re-allocating budget, with SSB-CE again providing the best results.

\section{Conclusion}
\label{sec:conclusion}

This article is motivated by literature that recognises (1) the difficulty of recognising easy instances~\cite{figueira2017easy}; (2) the computational cost of deriving information needed to train algorithm-selectors; (3) the fact that algorithm selectors do not perform well on easy instances~\cite{kerschke_automated_2019}; (4) the need to make the most efficient use of a fixed function evaluation budget in an algorithm-selection pipeline. 
To address these issues, we proposed a pipeline (Fig.~\ref{fig:pipeline}) that includes a `hardness' classifier whose role is to filter out easy instances which are simply solved using the SBS.
The effect of including this classifier is two-fold: (1) it \textit{saves} budget both by removing the need to calculate some of the input required by an algorithm-selector and by curtailing the run length of easy instances; (2) it \textit{re-allocates} this budget to hard instances, enabling larger solving budgets which results in better performance.

The pipeline was evaluated over two scenarios found in practical applications, i.e., a batch setting where all instances to be solved are known, and a streaming setting in which instances arrive one at time and must be solved immediately.

Over a stream of instances generated from the BBOB benchmark suite, we show that the proposed pipeline including re-allocation results in a gain in performance compared to using a trained algorithm-selector on the hard instances.
The magnitude of the gain increases as the amount of budget saved and re-allocated increases.
We also show that we can improve on an `oracle' which uses a fixed budget to identify the VBS (albeit unrealistic in practice). This is possible due to the dynamic re-allocation of saved budget which increases run length --- using the pipeline results in $7\%$ of budget becoming available for re-allocation.

A critical aspect of the design of the pipeline is that there should not be a cost to computing the data required as input to the hardness classifier which would negate the proposed benefits.
We addressed this by using a hardness classifier whose input is a \textit{subset} of the data required by the algorithm-selector, therefore needs to be computed regardless of whether or not the hardness classifier is used. 
Specifically, we use short fitness trajectories as input, which fit the criterion identified above and have recently been shown to outperform selectors based on ELA features~\cite{RenauH24}.
Note that a pipeline that used ELA features as input to both classifiers would in fact also fit this criterion as feature vectors calculated for the hardness classifier could directly be used in the algorithm-selector.
However, although a pipeline using ELA features could achieve some performance benefits, budget would only be saved by identifying easy instances and reducing the length of the run of the SBS solver (as opposed to the trajectory approach which also saves budget by reducing feature computation). In addition, in some circumstances, incorrectly identifying instances as easy and solving with the SBS can mitigate an incorrect prediction with the algorithm-selector.

Future work can be separated into two categories: (1) improving the hardness classifier and (2) improving the strategy for re-allocating budget. 
The accuracy of the hardness classifier could be improved by addressing the imbalance in the training dataset, for example generating new easy instances through instance-generation methods (e.g., ~\cite{munoz2020generating}) or using imbalance correction techniques such as weighting, down-sampling, or over-sampling.  
With regard to budget re-allocation, we used a na\"{i}ve strategy of spending for both the batch and streaming scenarios.
Alternative methods could be explored such as spreading the budget over the next $h$ `hard' instances in the stream or even predicting the level of `hardness' via regression rather than a binary label of easy/hard.

\begin{credits}
\subsubsection{\ackname} The authors are supported by funding from EPSRC award number: EP/V026534/1

\clearpage
\subsubsection{\discintname}
The authors have no competing interests to declare that are
relevant to the content of this article. 
\end{credits}

\bibliographystyle{splncs04}
\bibliography{references}
\end{document}